\documentclass[journal]{IEEEtran}
\usepackage{amsmath}
\usepackage{float}
\usepackage{booktabs}
\usepackage{multirow}
\usepackage{hyperref}
\usepackage{graphicx}
\usepackage{amsmath,amssymb}
\ifCLASSINFOpdf

\else

\fi

\begin{document}
	
	\title{MT-RGCN-DTI: A Multi-Task Relational Graph Convolutional Network for Drug–Target Interaction Prediction}
	
	\author{Dheeran Sankaran\textsuperscript{1}, 
		R Yadunandan\textsuperscript{1} %
		Daniele Santoni\textsuperscript{2}
		and Nimisha Ghosh\textsuperscript{1, *}%
		\thanks{\textsuperscript{1}Department of Computer Science and Engineering, Shiv Nadar University Chennai, Tamil Nadu, India}%
		\thanks{\textsuperscript{2}Institute for System Analysis and Computer Science “Antonio Ruberti”, National Research Council of Italy, Rome, Italy}%
		\thanks{*Corresponding author: Nimisha Ghosh (nimishaghosh@snuchennai.edu.in).}}
	\maketitle
	
	\begin{abstract}
		Prediction of drug-target interaction (DTI) has become a very crucial part of drug discovery. With the increase in the data volume, it has become very important to improve prediction accuracy by integrating multiple data sources. In this regard, utilising multiple resources in the form of heterogeneous information network to address DTI prediction can be rewarding. In this work, we propose a heterogeneous information network based multi-task DTI prediction model which leverages the advantages of relation graph convolutional network (RGCN). To achieve this, the model employs drug-target interaction network, drug-drug similarity network and protein-protein similarity network to construct a heterogeneous information network. Moreover, we use a knowledge  graph embedding scoring functions to model interactions between drug and targets, specifically DistMult and ComplEx as decoders over the learned node representations. We performed extensive experiments on multiple benchmark datasets and compared with state-of-the-art methods. We observed that the proposed model shows competitive performance, thereby confirming the effectiveness of our approach. The codes and datasets are available at https://github.com/Dheeran-S/MT-RGCN-DTI/.
	\end{abstract}
	
	
	\IEEEpeerreviewmaketitle

	\section{Introduction}
	Drug discovery is a cumbersome task which needs substantial amount of funding to develop new drugs catering to various number of diseases. However, such newly discovered drugs may fail to get proper approval for clinical use due to many factors such as unexpected drug reactions and low binding affinity~\cite{Lee2024}. In this regard, artificial intelligence has turned out to be quite a promising tool for reducing high expenses in various facets of drug discovery. The first and foremost step in this direction is drug-target interaction (DTI) predictions. In the last few years, significant amount of progress has been made in this direction resulting in improved drug discovery techniques. This includes both machine learning (ML)~\cite{Laarhoven2011, Nascimento2016, Ding2014} and network-based~\cite{Luo2017, Chen2012, Cheng2012} methods. In ML-based methods, the drugs and target proteins are mostly encoded using molecular descriptors to predict interactions while network-based methods rely on interaction network between drugs and targets for such prediction. In~\cite{Debnath2025, AHMAD2026}, Debnath et al. and Ahmad et al. have reviewed the latest progress in the domain of drug-target interactions. These works provide a detailed overview of the different models that have been used in the recent past for DTI predictions. 
	
	One of the most popular methods in DTI is Graph Neural Networks (GNNs) that have shown commendable results. The main idea behind such methodology is node and graph representation learning using node attributes and graph representation. Recently, WDGBANDTI~\cite{Wang2025} proposed by Wang et al. uses deep graph convolutional network (GCN), convolutional neural networks (CNN) and a bilinear attention network for DTI prediction. Wan et al. proposed NeoDTI~\cite{Wan2019} that utilises GCNs to apply to several relational networks to learn the representation of biological entity nodes. DeepMGT-DTI~\cite{ZHANG2022} uses multilayer graph fusion approach using a transformer network to collect information from different layers of GCN. To enhance edge feature learning, they also implemented a multi-headed attention architecture. GCN-DTI~\cite{Zhao2021} uses GCN to learn the characteristics of a drug-protein pair network formed using multiple drugs and proteins as inputs. Such learned feature representation are then fed into a deep neural network for the final DTI prediction. In NHGNN-DTA~\cite{He2023}, He et al. use a hybrid GNN that sequence-based feature generation with graph-level information interaction that include a unique central node connecting drug and protein graphs. For cold-start scenarios, they used cold-drug, cold-target and cold-drug-target. For interpretatbility, they used attention-weight visualisation of the final output. Wang et al. proposed MSN-DTA~\cite{WANG2025_DTA} that uses multi-scale graph isomorphic network to extract multi-level structural information. Here also, to aid in interpretability, attention-weight visualisation is used for the protein binding sites. GS-DTA~\cite{Luo2025} proposed by Luo et al. utilises the benefits of GATv2-GCN and three-layer GCN for drug-feature extraction while CNN, Bidirectional Long Short-Term Memory (BiLSTM) and Transformer are used for protein feature extraction for prediction drug-target binding affinity (DTA). A new multiview graph convolutional network (MVGCN) framework is proposed in~\cite{Fu2022} for link prediction in biomedical binary networks.
	
	Recently, integration of multiple data sources and information have gained popularity in DTI prediction. In this regard, DTINet~\cite{Luo2017} automatically learns low-dimensional feature representations of drugs and targets from heterogeneous data and then use inductive matrix completion to predict DTI. The drawback of this work is that it focusses only on feature extraction of drugs and proteins but not on their interaction information, thereby losing important insight. To somewhat mitigate this issue, NormMulInf~\cite{Peng2017} proposed a semisupervised model both with and without labels that combines PCA and various biological information. However, based on its dependency on semisupervised learning framework and biological information, it lacks the ability to handle complex drugtarget relationships as well as multitask learning. To mitigate such limitations, heterogeneous information network (HIN) methods have been put forth for DTI prediction. HIN has the capability to combine different types of data that include drug chemical structure, protein sequence and gene ontology (GO) annotations, thereby providing a more comprehensive insight of the relationships between drugs and targets. Such methodology has been adopted in~\cite{Wang2019} where the authors have proposed a heterogeneous graph attention network viz. HAN that uses node- and semantic-level attention, thereby allowing the model to consider the benefits of node as well as meta-paths and enhancing DTI predictions. The idea of heterogeneous graph is further explored in~\cite{Deng2025} where Deng et al. propose a multi-task DTI prediction approach (HIN-MTDTI) using a unified heterogeneous information network (HIN) that integrates information from drug-target interaction network, drug-drug similarity network and target-target similarity network. Subsequently, to learn the drug and target representations and perform DTI prediction, HIN-MTDTI employs GCN. Furthermore, the model also uses bilinear attention network to capture local drug-target interactions. The authors perform multi-task learning in the form of classification for DTI and regression for DTA. 
	
	Taking cues from the literature, in this work we propose MT-RGCN-DTI which is a heterogeneous graph-based framework that leverages a Relational Graph Convolutional Network (RGCN)~\cite{Schlichtkrull2023} for multi-task learning. Similar to HIN-MTDTI~\cite{Deng2025}, this work also integrates multiple biological networks such as drug-target interaction network, drug-drug similarity network and protein-protein similarity network. However, unlike~\cite{Deng2025} which treats the auxiliary relations like drug-drug and protein-protein as a single graph along with drug-target, our model explicitly represents them as distinct relation types. This allows the model to fully exploit the unique characteristics of each interaction type through relation-specific transformations. This is important as we use knowledge graph embedding scoring functions like DistMult and ComplEX to model the interactions between drug and targets. To enhance representation quality, we introduce an attention-based gating mechanism that selectively emphasizes informative embedding dimensions prior to interaction scoring. Extensive experiments on multiple benchmark datasets demonstrate that the proposed framework consistently outperforms state-of-the-art methods for both drug–target interaction (DTI) classification and drug–target affinity (DTA) regression tasks.
	
	To summarise, the contributions of this work are as follows:
	\begin{itemize}
		\item We propose a multi-relational graph learning framework that integrates heterogeneous biological interactions (DTI, DDI, and PPI) using RGCN, enabling relation-specific representation learning.
		\item We introduce an attention-based embedding refinement mechanism that enhances interaction-relevant features prior to prediction.
		\item We formulate drug-target interaction prediction as a multi-task learning problem, where DDI and PPI prediction act as auxiliary tasks to learn richer graph representations while improving the primary DTI prediction task for both classification and regression.
		\item We design two knowledge graph embedding-based decoders built upon DistMult and ComplEx scoring functions, augmented with a nonlinear multilayer perceptron (MLP) to capture higher-order interaction patterns beyond conventional bilinear scoring.
		
	\end{itemize}
	
	\section{Materials and Methods}
	In this section, we describe the datasets used in this work followed by the methodology used. 
	\subsection{Datasets}
	For the DTI prediction, we use Enzyme and IC datasets from~\cite{Yamanishi2008} and DrugBank dataset\footnote{https://go.drugbank.com/releases/latest}. The datasets from~\cite{Yamanishi2008} include DTI interactions in humans involving enzymes and ion channels while those collected from DrugBank contains all FDA approved drugs with the corresponding protein targets. On the other hand, for the DTA prediction, we use DAVIS AND KIBA datasets. Both of these are standard datasets used for prediction of affinity. The Davis dataset contains specific tests for
	kinase protein families and the associated inhibitors' dissociation constants ($K_d$). The KIBA dataset combines the biological activity of kinease inhibitors from different sources like $K_i$, $K_d$ and $IC_{50}$. Table~\ref{tab1} reports the dataset details.
	\begin{table}[H]\scriptsize
		\centering
		\caption{Summary of Data}
		\label{tab1}
		\begin{tabular}{lllll}
			\hline
			Dataset & Task  & Drug  & Protein & Interactions \\
			\hline
			EN & Classification & 445 & 663 & 2925\\
			IC & Classification & 210 & 204 & 1476\\
			DrugBank & Classification & 1555 & 1593 & 5837\\
			Davis & Regression & 68 & 379 &  25772\\
			KIBA & Regression & 2068 & 229 & 117657\\\hline
		\end{tabular}
		
	\end{table}
	\subsection{Methodology}
\begin{figure*}
		\centering
		\includegraphics[height=3.8in,width=5.0in]{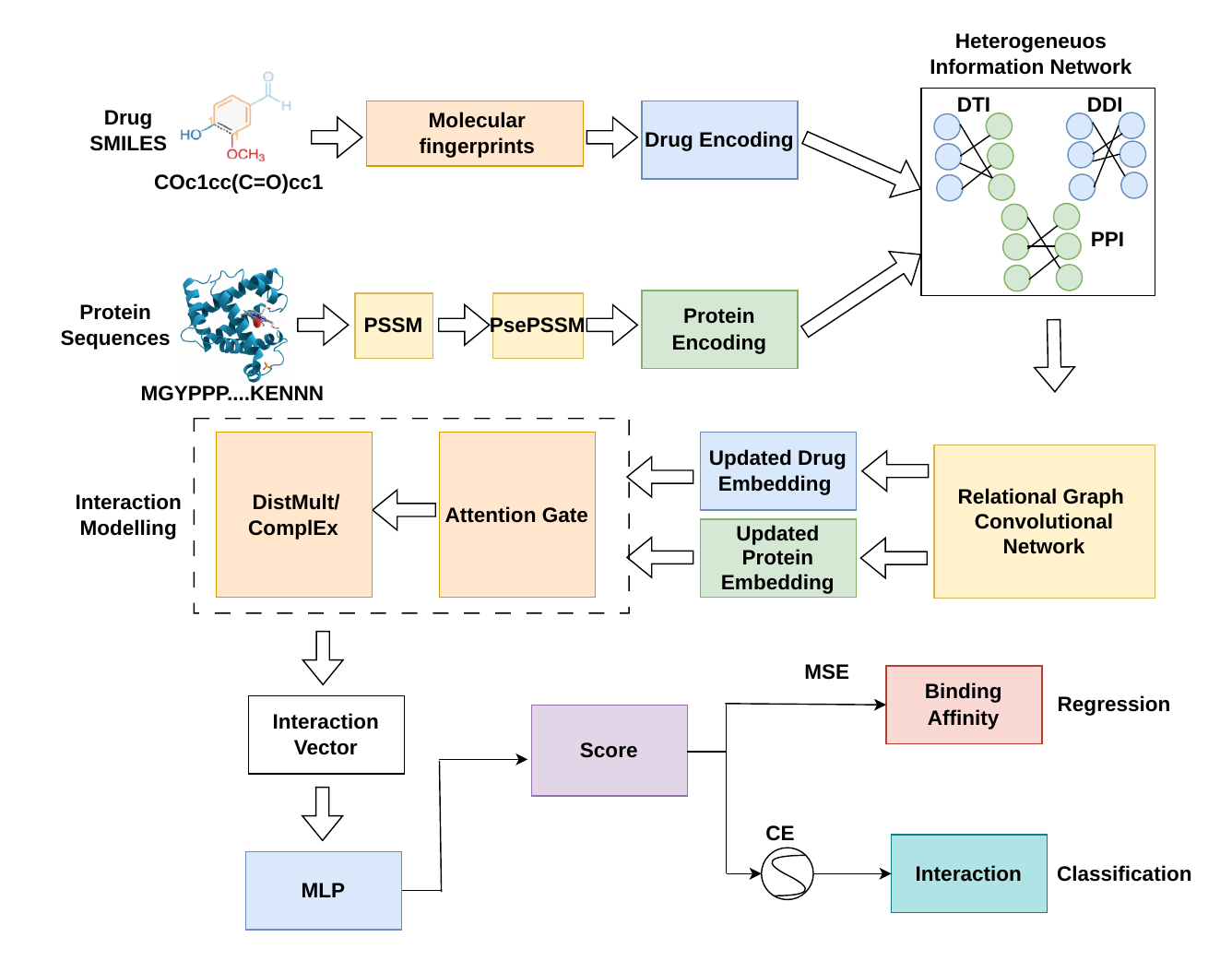}
		\caption{Pipeline of the work}
		\label{fig:pipeline}
	\end{figure*}

	The proposed model consists of separate feature encoders for drugs and proteins, a graph embedding module based on a Relational Graph Convolutional Network (RGCN), a dimension-wise attention gating module and a decoder for interaction prediction. The decoder employs DistMult or ComplEx knowledge graph embedding functions to construct an interaction vector, which is subsequently processed by a multilayer perceptron (MLP) to produce a final interaction score. For classification, a sigmoid activation converts this score into an interaction probability, whereas for regression the output directly represents the predicted binding affinity. Figure~\ref{fig:pipeline} shows the overall workflow of the proposed model.
	\subsubsection{Drug Feature Representation}
	Molecular fingerprints are fixed-length vector representations of chemical compounds that encode the structural and chemical properties of molecules. These representations enable efficient computation and serve as input features for learning models. For encoding drugs. molecular fingerprints are widely used as they can transform molecular structures into such form of vectors as understandable by the models. Following~\cite{Deng2025}, we have used the chemical substructure fingerprint as given in the PubChem database. Each drug molecule is converted into a binary vector:
	$\mathbf{x}_d = [b_1, b_2, \dots, b_d]$, where $b_i \in \{0,1\}$ indicates the presence or frequency of a substructure.

	\subsubsection{Protein Feature Representation}
	In this work, protein sequences are represented by pseudo position specific scoring matrix (PsePSSM) features. PsePSSM enables encoding of protein sequences of varying lengths. In this regard, for a protein sequence $P_n$ of length $l$, a position specific scoring matrix (PSSM) of dimensions $l \times 20$ is generated (Here, 20 is the number of amino acids):
	\begin{equation}
		\mathrm{PSSM}(P_n) =
		\begin{bmatrix}
			E_{1 \rightarrow 1} & E_{1 \rightarrow 2} & \cdots & E_{1 \rightarrow 20} \\
			E_{2 \rightarrow 1} & E_{2 \rightarrow 2} & \cdots & E_{2 \rightarrow 20} \\
			\vdots & \vdots & \ddots & \vdots \\
			E_{l \rightarrow 1} & E_{l \rightarrow 2} & \cdots & E_{l \rightarrow 20}
		\end{bmatrix}
	\end{equation}
	In order to create the matrix, the initial step is to find homologous sequences using tools such as PSI-BLAST. Once such sequences are identified, multiple sequence alignment (MSA) techniques like ClustalW are applied to align similar sequences. Thereafter, for each position $i$ in the alignment, the frequency of each amino acid is calculated which is then converted to a score based on the replacement of amino acid $i$ with the standard amino acid $j$.  This score is represented by $E_{i \rightarrow j}$ in the matrix which is then normalised within the (0, 1) range using the sigmoid function:
	\begin{equation}
		\bar{E}_{i \rightarrow j} = \frac{1}{1+exp(E_{i \rightarrow j})}
	\end{equation}
	Subsequently, the average score of amino acid $j$ across a sequence of length $l$ is given by:
	\begin{equation}
		\bar{E_j} = \frac{1}{l}\sum_{i=1}^{l} \bar{E}_{i \rightarrow j}
	\end{equation}
	Finally, the pseudo position specific scoring matrix (PsePSSM) for protein $\mathcal{P}$ is calculated as:
	\begin{equation}
		\begin{split}
			PsePSSM^\lambda_{\mathcal{P}}{(p_n)} & = \\ [\bar{E_1},\bar{E_2},\dots\bar{E_n},\theta_1^1,\theta_2^1,\dots\theta_{20}^1,\theta_1^\lambda,\theta_2^\lambda,\dots\theta_{20}^\lambda]^T
		\end{split}
	\end{equation}
	Here, $T$ denotes the transpose operator and $\theta_j^\lambda = \frac{1}{l-\lambda}\sum_{i=1}^{l-\lambda}[\bar{E}_{i \rightarrow j}-\bar{E}_{(i+\lambda) \rightarrow j}]^2$ and $j=1,\dots,20, 0 \leq \lambda \leq l$. Here, $\lambda$ shows how far apart positions are compared while $\theta_j^\lambda$ shows how much the importance of $j^{th}$ amino acid changes across that distance and $l$ is the length of a protein sequence.  
	
	\subsection{Drug-Target Heterogeneous Information Network}
	In order to model the complex relationships between drugs and proteins, we construct a heterogeneous information network (HIN) that combines multiple biological interaction types such as Drug-Target Interaction (DTI),  Drug-Drug Interaction (DDI) and Protein-Protein Interaction (PPI). This network is defined as multi-relational graph $\mathcal{G} = (\mathcal{V}, \mathcal{E}, \mathcal{R})$, where $\mathcal{V}$ denotes the set of nodes, $\mathcal{E}$ represents edges and $\mathcal{R}$ is the set of relation types. In this regard, the node consists of two distinct types: Drug Nodes ($\mathcal{V}_d$) and Protein or Target Nodes ($\mathcal{V}_p$). Thus, $\mathcal{V} = \mathcal{V}_d \cup \mathcal{V}_p$. On the other hand, we have three biologically meaningful relations as DTI, DDI and PPI. Formally, $\mathcal{R} = \{ r_{DTI}, r_{DDI}, r_{PPI} \}$. Each relation $r \in \mathcal{R}$ is associated with an adjacency matrix $\mathbf{A}_r$. Thus, the heterogeneous graph can be expressed as a collection of relation-specific adjacency matrices $\{\mathbf{A}_{DTI}, \mathbf{A}_{DDI}, \mathbf{A}_{PPI}\}$, where each matrix encodes connectivity under a specific relation type, thereby ensuring relation-aware message passing in the subsequent RGCN layers.
	
	The initial graph is constructed from known DTI interactions and subsequently enriched with DDI and PPI similarity edges. For DTI, drug nodes are represented using molecular fingerprints while protein nodes are represented using sequence-derived embeddings based on PsePSSM-based features. Let  $\mathbf{X}_d$ $\in$ $\mathbb{R}^{N_d \times d_d}$ and $\mathbf{X}_p \in \mathbb{R}^{N_p \times d_p}$ denote the feature matrices for drugs and proteins, respectively. These features are first transformed into a shared latent space using separate feature encoders:
	\begin{equation}
		\mathbf{H}^{(0)} = \text{LN}(\text{BN}([\text{Enc}_p(\mathbf{X}_p);\ \text{Enc}_d(\mathbf{X}_d)]))
	\end{equation}
	where each encoder is defined as:
	\begin{equation}
		\text{Enc}(x) = \text{Dropout}(\text{LN}(\text{BN}(\text{ReLU}(Wx))))
	\end{equation}
	Here, $\mathbf{H}^{(0)}$ is the initial node embedding matrix representing all node representations (drugs + proteins) in the graph and and $W_p$ and $W_d$ (given as $W$) are learnable weight matrices corresponding to protein and drug feature transformations respectively. $LN$ represents Layer Normalization and $BN$ is Batch Normalization. This transformation ensures that both fingerprint and PsePSSM lie in the same latent space before graph propagation. Once the unified node representations $\mathbf{H}^{(0)}$ are obtained, we construct similarity-based relations to enrich the network structure. Specifically, a drug–drug similarity network is computed using Tanimoto coefficient over molecular fingerprint features, while a protein–protein similarity network is constructed using cosine similarity over PsePSSM-based features. Thus, $\mathbf{A}_{DDI}(i,j) = \text{Tanimoto}(d_i,d_j)$ and $\mathbf{A}_{PPI}(i,j) = \text{cos}(p_i,p_j)$. These similarity measures define additional edge sets $\mathcal{E}_{DDI}$ and $\mathcal{E}_{PPI}$, which are incorporated into the heterogeneous graph alongside the original DTI edges. 
	
	\subsection{Relational Graph Convolutional Network}
	With the obtained multi-relational graph representing DTI, DDI and PPI, we use a Relational Graph Convolutional Network (RGCN) to learn node embeddings. Using RGCN, the node representations (embeddings) are updated as:
	\begin{equation}
		\mathbf{h_i}^{l+1} = \sigma \left(\sum_{r \in \mathcal{R}} \sum_{j \in \mathcal{N}_i^r}\frac{1}{c_{i,r}}\mathbf{W}_r^{(l)} \mathbf{h}_j^{(l)}+\mathbf{W}_0^{(l)} \mathbf{h}_i^{(l)}
		\right)
	\end{equation}
	This equation describes how the representation of node $i$ is updated at layer $(l+1)$ in RGCN. Here, $\mathbf{h_i}^{(l)}$ is the embedding at layer $l$ while $\mathbf{h_i}^{(l+1)}$ is the updated embedding at the next layer. $\mathcal{R}$ is the set of relation types such as DTI, DDI and PPI ($r$), $\mathcal{N}_i^r$ represents the neighbours of node $i$ for relation $r$, $\mathbf{W}_r^{(l)}$ and $\mathbf{W}_0^{(l)}$ is the weight matrix specific to relation $r$ and the node's own features (self-loop) respectively. $\frac{1}{c_{i,r}}$ is the normalisation factor. Finally, $\sigma(.)$ is the activation function. After $L$ layers, the final node representation is given as:
	\begin{equation}
		\mathbf{H}^{(L)} = \text{RGCN}(\mathbf{H}^{(0)})
	\end{equation}
	
	\subsection{Interaction Modelling}
	After the final representation of drugs and proteins are obtained ($\mathbf{H}^{(L)}$),  such learned embeddings are used to predict different biological relations viz. DTI, DDI and PPI. Each relation is modeled using its own relation embedding while sharing the same graph encoder, allowing the auxiliary DDI and PPI objectives to improve the learned node representations.
	Here, the goal is to estimate the likelihood of interaction of two nodes based on their updated learned embeddings. Before computing interaction scores, the node embeddings are refined using a dimension-wise attention gating mechanism thus effectively silencing irrelevant dimensions of the embedding vector and amplifying important ones before deciding the final score. For a given pair of nodes, let the corresponding embeddings for drug and protein extracted from the final matrix be $\mathbf{h} = \mathbf{H}^{(L)}[i]$ and $\mathbf{t} = \mathbf{H}^{(L)}[j]$ denote the embeddings of the head and tail entities, respectively. The attention weights are computed as:
	\begin{equation}
		\boldsymbol{\alpha} = \sigma_{\text{att}}(\mathbf{W}_h \mathbf{h} + \mathbf{W}_t \mathbf{t} + \mathbf{b})  
	\end{equation}
	The refined (gated) embeddings are then obtained as:
	\begin{equation}
		\tilde{\mathbf{h}} = \boldsymbol{\alpha} \odot \mathbf{h}, \quad
		\tilde{\mathbf{t}} = \boldsymbol{\alpha} \odot \mathbf{t} 
	\end{equation}
	These attention-weighted embeddings emphasize interaction-relevant feature dimensions and are subsequently used for interaction modelling.
	
	For any two given nodes, such embeddings are combined using a bilinear scoring function in the normal or complex space to produce an interaction score; specifically, we use knowledge-graph based scoring function such as DistMult and ComplEx for this purpose. However, instead of directly summing the interaction vector as in the original DistMult and ComplEx formulations, the proposed framework preserves the interaction vector and feeds it to a three-layer Multilayer Perceptron (MLP), enabling nonlinear modelling of higher-order interactions.
	These approaches are considered as they can be modelled for triplets like $(h,r,t)$ where $h$ and $t$ respectively represent head and tail entities (nodes which can be drug or protein) while $r$ represents their relationship (DTI, DDI or PPI). Thus, such approaches capture the compatibility between node embeddings with respect to a given relation while maintaining computational efficiency. DistMult captures interactions through element-wise multiplication of embeddings. However, DistMult works on the assumption that relations are inherently symmetric and is therefore naturally suited to modelling symmetric similarity relations, whereas ComplEx additionally captures asymmetric interactions such as DTI. By introducing real and imaginary components, ComplEx can model asymmetric interactions effectively. To fully utilise the benefits of both these scoring functions, the model incorporates relation-specific transformations and attention mechanisms that allow it to distinguish between different types of interactions. This is particularly important in heterogeneous graphs, where each edge type carries distinct biological meaning. By jointly modeling multiple interaction types, the framework captures complementary information across different biological networks, leading to more robust and biologically meaningful predictions.
	DistMult ($v_{DM}$) and ComplEX ($v_{CX}$) (jointly denoted as $\mathbf{v}$),  that can be expressed as: 
	\begin{equation}
		\mathbf{v_{DM}}(h,r,t) = \tilde{h}_h \odot r \odot \tilde{h}_t
	\end{equation}
	Here, $\tilde{h}_h$ and $\tilde{h}_t$ are the embedding vectors of the head and tail entities (drug and proteins).
	Unlike DistMult, ComplEx represents each embedding as a complex-valued vector comprising real and imaginary components, i.e., $\mathbf{h}=\mathbf{h}_r+i\mathbf{h}_i$, $\mathbf{r}=\mathbf{r}_r+i\mathbf{r}_i$, and $\mathbf{t}=\mathbf{t}_r+i\mathbf{t}_i$. The interaction vector is obtained by computing the real part of the complex product $\operatorname{Re}(\mathbf{h}\odot\mathbf{r}\odot\overline{\mathbf{t}})$, where $\overline{\mathbf{t}}$ denotes the complex conjugate of the tail embedding. Expanding this product yields the following interaction vector:
	\begin{equation}
		\begin{aligned}
			\mathbf{v_{CX}}(h,r,t) = \; &
			\tilde{h}_{h_r} \odot r_r \odot \tilde{h}_{t_r} + \tilde{h}_{h_r} \odot r_i \odot \tilde{h}_{t_i} \\ 
			&+ \tilde{h}_{h_i} \odot r_r \odot \tilde{h}_{t_i} - 
			\tilde{h}_{h_i} \odot r_i \odot \tilde{h}_{t_r}
		\end{aligned}
	\end{equation}
	Here, the suffices $r$ and $i$ denote the real and imaginary part.
	Although the interaction vector $\mathbf{v}$ is computed for a single relation type, the underlying node embeddings have already incorporated structural information propagated through DTI, DDI and PPI edges. Consequently, each prediction benefits from knowledge learned across multiple biological interaction networks.
	However, the final prediction task focuses exclusively on DTI. The auxiliary DDI and PPI tasks provide additional structural supervision, enabling the model to learn more informative embeddings, but only the DTI scores are used during inference and evaluation.
	
	\subsection{Drug-Target Interaction Prediction}
	Although, we have used DTI, DDI and PPI during training, the primary objective of this work is to predict DTI using the learned node representations. In this regard, we have designed a multi-task learning framework that supports both DTI classification and DTA regression by employing task-specific prediction heads and loss functions. To achieve this, we process the previously $\mathbf{v}$ through a 3-layer Multi-Layer Perceptron (MLP) to generate an interaction score:
	\begin{equation}
		z_{ij} = MLP(\mathbf{v}) + b
	\end{equation}
	For the classification part, the output of this MLP is converted into a probability using a sigmoid function:
	\begin{equation}
		\hat{y}_{ij} = Sigmoid(MLP(\mathbf{v}) + b)
	\end{equation}
	The corresponding binary cross-entropy loss is calculated using:
	\begin{equation}
		\mathcal{L_{CE}} = -\frac{1}{N}\sum_{i=1}^N[{y}_{ij}log(\hat{y}_{ij}) + (1-{y}_{ij})log(1-\hat{y}_{ij})]  
	\end{equation}
	where, ${y}_{ij}$ denotes the ground truth. For the regression task, this continuous score $z_{ij}$ is directly used as the predicted binding affinity ($\hat{z}_{ij} = z_{ij}$). The model is optimized using Mean Squared Error (MSE):
	\begin{equation}
		\mathcal{L_{MSE}} = \frac{1}{n}\sum_{i=1}^n({y}_{ij} - \hat{z}_{ij})^2
	\end{equation}
	where $y_{ij}$ denotes the ground truth continuous affinity.
	Similar equations are also used on ComplEX score for classification and regression tasks. 
	
	Thus, the multi-task optimisation can be summed up as:
	\begin{equation}
		\mathcal{L} = \mathcal{L_{DTI/DTA}} + \lambda_{DDI}\mathcal{L_{DDI}} + \lambda_{PPI}\mathcal{L_{PPI}}
	\end{equation}
	where, $\lambda$ is the fixed weighting coefficient.
	\section{Results and Discussions}
	This section discusses the results as obtained from this work along with the experimental setup. We have reported the results for both DistMult and ComplEX for classification and regression tasks.
	\subsection{Experimental Setup}
	For our experiments, we have employed similar setup as~\cite{Deng2025}. The input dimensions of PsePSSM is 220 while that of fingerprint is 881 and the output dimension is 256. The hidden layer dimension of RGCN is 256 and the corresponding number of graph layers is 2.  The training runs for 1500 epochs with an early stopping criteria of 20 epochs. The learning rate is fixed at 0.0005, the weight decay is set to 5e-4 while the dropout rate is fixed at 0.3. We use five-fold cross-validation for reporting the final results. The final prediction result of DTI is based on the information from DDI and PPI as well. In this regard, the fixed weighting coefficient ($\lambda$) of $\mathbf{A}_{DDI}$ and $\mathbf{A}_{PPI}$ are each considered as 0.3 while the minimum similarity score required to create a graph edge between two drugs or two targets is set at 0.5.

	\subsection{Evaluation Metrics}
	To judge the performance of our model, we have used different standard metrics such as Accuracy, Precision, Sensitivity, Specificity, Matthews Correlation Coefficient (MCC), Area Under Curve (AUC) and Area Under the Precision-Recall Curve (AUPR) for classification purposes while for regression, metrics such as  Mean Squared Error (MSE) and Concordance Index (CI) are considered. CI is used to calculate difference between the predicted and true values. It is calculated as:
	\begin{equation}
		CI = \frac{1}{Z}\sum_{d_x > d_y} h(b_x - b_y)
	\end{equation}
	where, $b_x$ and $b_y$ are the predicted values of larger and smaller affinities $d_x$ and $d_y$ respectively while $Z$ is the normalisation constant and $h(x)$ is the Leap function.
	\subsection{Comparison with existing method}
	To demonstrate the efficiency of MT-RGCN-DTI, we compare our proposed model with state-of-the-art methods, HIN-MTDTI~\cite{Deng2025}. As, HIN-MTDTI model is published recently and it has already shown performance improvement as compared to other methods such as~\cite{Sajadi2021, Peng2017, Bian2023, Pahikkala2015, Oz2018} etc., HIN-MTDTI is chosen as the benchmark model. It uses molecular fingerprints and PsePSSM for drug and protein feature representations respectively. Graph Convolutional Network (GCN) is subsequently used for learning such representations and predict their interactions. Furthermore, bilinear attention network is used to capture local drug-target interaction information. They have also performed multitask learning for classification in the form of DTI as well as regression task for drug-target binding affinity. The results of the classification task are reported in Table~\ref{tab:combined_DTI} while the results of regression are reported in Table~\ref{tab:combined_DTA}. We have reported the results for both DistMult (MT-RGCN-DTI-DM) and ComplEX (MT-RGCN-DTI-CX).
	
	\subsubsection{Classification}
	As can be observed from the results of Table~\ref{tab:combined_DTI}, our proposed model shows competitive performance for all the reported metrics in DTI tasks. For enzyme dataset, the proposed model (MT-RGCN-DTI-DM/ MT-RGCN-DTI-CX) has Accuracy, Precision, Sensitivity, Specificity, MCC, AUC and AUPR of 0.919, 0.936, 0.906, 0.939, 0.839, 0.974 and 0.978 respectively.
	Thus, the proposed model shows improvements in Accuracy, Sensitivity, MCC, AUC and AUPR by 0.3\%, 5.47\%, 0.1\%, 0.9\% and 0.5\% respectively. MT-RGCN-DTI-CX performs better for all the reported metrics except Sensitivity where  MT-RGCN-DTI-DM performs better. 
	For IC dataset, the best performance is achieved by the proposed model for all the metrics except Precision with the values being 0.990 (Accuracy),  0.785 (Sensitivity), 0.836 (MCC), 0.974 (AUC) and 0.877 (AUPR), thus showing improvements of  0.01\%, 0.3\%, 0.2\%, 6.8\%, and 8.67\% respectively, while for specificity, both HIN-MTDTI and MT-RGCN-DTI show the same value of 0.998. However, for the DrugBank dataset, HIN-MTDTI show better results for most of the parameters with similar results for AUC and AUPR. 
	
	\begin{table*}
		\centering
		\caption{Performance comparison of MT-RGCN-DTI with existing state-of-the-art across Enzyme, IC and DrugBank datasets for classification tasks}
		\label{tab:combined_DTI}
		\scriptsize
		\begin{tabular}{lcccccccc}
			\toprule
			\textbf{Dataset} & \textbf{Method} & \textbf{Acc} & \textbf{Prec} & \textbf{Sen} & \textbf{Spec} & \textbf{MCC} & \textbf{AUC} & \textbf{AUPR}\\
			\midrule
			
			\multirow{3}{*}{Enzyme} 
			& HIN-MTDTI & 0.916 $\pm$ 0.016 & \textbf{0.970 $\pm$ 0.007}
			& 0.859 $\pm$ 0.029 & \textbf{0.974 $\pm$ 0.006} & 0.838 $\pm$ 0.029 & 0.965 $\pm$ 0.007 & 0.973 $\pm$ 0.005 \\
			& MT-RGCN-DTI-DM & 0.914 $\pm$ 0.008 & 0.920 $\pm$ 0.012 & \textbf{0.906} $\pm$ 0.020 & 0.921 $\pm$ 0.014 & 0.827 $\pm$ 0.017 & 0.968 $\pm$ 0.006 & 0.972 $\pm$ 0.005 \\
			& MT-RGCN-DTI-CX & \textbf{0.919 $\pm$ 0.004} & 0.936 $\pm$ 0.008 & 0.899 $\pm$ 0.015 & 0.939 $\pm$ 0.010 & \textbf{0.839 $\pm$ 0.008} & \textbf{0.974 $\pm$ 0.005} & \textbf{0.978 $\pm$ 0.004} \\
			
			\midrule
			\multirow{3}{*}{IC}
			& HIN-MTDTI & 0.989 $\pm$ 0.001 & \textbf{0.949 $\pm$ 0.016} & 0.782 $\pm$ 0.010 & \textbf{0.998 $\pm$ 0.000} & 0.834 $\pm$ 0.017 & 0.912 $\pm$ 0.010 & 0.807 $\pm$ 0.017 \\
			& MT-RGCN-DTI-DM & 0.989 $\pm$ 0.002 & 0.926 $\pm$ 0.016 & 0.746 $\pm$ 0.031 & \textbf{0.998 $\pm$ 0.001} & 0.826 $\pm$ 0.021 & \textbf{0.974} $\pm$ 0.005 & 0.863 $\pm$ 0.023 \\
			& MT-RGCN-DTI-CX & \textbf{0.990} $\pm$ 0.003 & 0.903 $\pm$ 0.024 & \textbf{0.785} $\pm$ 0.052 & 0.997 $\pm$ 0.001 & \textbf{0.836} $\pm$ 0.039 & 0.973 $\pm$ 0.008 & \textbf{0.877} $\pm$ 0.031 \\
			
			\midrule
			\multirow{3}{*}{DrugBank}
			& HIN-MTDTI & \textbf{0.874 $\pm$ 0.011} & \textbf{0.937 $\pm$ 0.006} & \textbf{0.801 $\pm$ 0.029} & \textbf{0.946 $\pm$ 0.007} & \textbf{0.756 $\pm$ 0.017} & \textbf{0.923 $\pm$ 0.008} & \textbf{0.931 $\pm$ 0.007} \\
			& MT-RGCN-DTI-DM & 0.850 $\pm$ 0.007 & 0.864 $\pm$ 0.012 & 0.831 $\pm$ 0.020 & 0.869 $\pm$ 0.016 & 0.701 $\pm$ 0.014 & 0.917 $\pm$ 0.006 & 0.926 $\pm$ 0.005 \\
			& MT-RGCN-DTI-CX & 0.849 $\pm$ 0.009 & 0.906  $\pm$ 0.012 & 0.780 $\pm$ 0.027 & 0.919 $\pm$ 0.014 & 0.706 $\pm$ 0.015 & \textbf{0.923 $\pm$ 0.006} & \textbf{0.931 $\pm$ 0.006} \\
			
			\bottomrule
		\end{tabular}
	\end{table*}
	\subsubsection{Regression}
	The regression results on two datasets, Davis and KIBA are reported in Table~\ref{tab:combined_DTA}. The results are based on MSE and CI. For Davis, both DM and CX surpass HIN-MTDTI in MSE and CI while for KIBA, DM shows better performance. MSE evaluates the accuracy of affinity value estimation, whereas CI evaluates the correctness of pairwise ranking. These metrics assess complementary aspects of regression performance. Based on the results in Table~\ref{tab:combined_DTA}, it can be seen that for both the datasets, the proposed method fares better. Considering the results from both Tables~\ref{tab:combined_DTI} and~\ref{tab:combined_DTA}, it can be observed that MT-RGCN-DTI achieves competitive and often superior performance across multiple benchmark datasets while providing a unified framework for both DTI classification and DTA regression. It is important to mention here that the authors of~\cite{Deng2025} have not released the DTA regression implementation; therefore, the released classification framework was adapted for affinity prediction.
	
	\begin{table}[H]
		\centering
		\caption{Performance comparison of MT-RGCN-DTI with existing methods across Davis and KIBA datasets for regression tasks}
		\label{tab:combined_DTA}
		\scriptsize
		\begin{tabular}{lccc}
			\toprule
			\textbf{Dataset} & \textbf{Method} & \textbf{MSE} & \textbf{CI} \\
			\midrule
			
			\multirow{3}{*}{Davis}
			& HIN-MTDTI & 0.228 $\pm$ 0.270 & 0.623 $\pm$ 0.007 \\
			& MT-RGCN-DTI-DM & 0.226 $\pm$ 0.016 & 0.876 $\pm$ 0.005 \\
			& MT-RGCN-DTI-CX & \textbf{0.213 $\pm$ 0.014} & \textbf{0.882 $\pm$ 0.003} \\
			\midrule
			
			\multirow{3}{*}{KIBA}
			& HIN-MTDTI & 0.317 $\pm$ 0.417 & 0.572 $\pm$ 0.011 \\
			& MT-RGCN-DTI-DM & \textbf{0.316 $\pm$ 0.010} & \textbf{0.796 $\pm$ 0.003} \\
			& MT-RGCN-DTI-CX & 0.379 $\pm$ 0.198 & 0.755 $\pm$ 0.112 \\
			\bottomrule
			
		\end{tabular}
	\end{table}
	
	\subsection{Ablation Studies}
	We have also performed ablation studies for both classification and regression tasks to showcase the effects of the absence and presence of each of the DDI and PPI matrices; Table~\ref{tab:ablation_DTI} and Figure~\ref{Ablation_figure} report such results. 
	\subsubsection{Classification}
	As reported in Table~\ref{tab:ablation_DTI}, for Enzyme dataset, the AUC values for DTI-only are 0.962 and 0.967 for DistMult and ComplEX respectively. Incorporating DDI increased AUC values to 0.964 and 0.970 respectively while with the inclusion of PPI, such values become 0.967 and 0.971. 
	\begin{table*}
		\centering
		\caption{Ablation Study of MT-RGCN-DTI in terms of AUC across Enzyme, IC and DrugBank datasets for classification tasks}
		\label{tab:ablation_DTI}
		\scriptsize
		\begin{tabular}{lccc}
			\toprule
			\textbf{Dataset} & \textbf{Model} & \textbf{MT-RGCN-DTI-DM} & \textbf{MT-RGCN-DTI-CX} \\
			\midrule
			\multirow{6}{*}{Enzyme} & DTI-only & 0.962 $\pm$ 0.009 & 0.967 $\pm$ 0.005\\
			& DTI + DDI & 0.964 $\pm$ 0.009 & 0.970 $\pm$ 0.007\\
			& DTI + PPI & 0.967 $\pm$ 0.006 & 0.971 $\pm$ 0.007\\
			& DTI + DDI ($\lambda$=0.5) + PPI ($\lambda$=0.5) & 0.966 $\pm$ 0.006 & 0.972 $\pm$ 0.005\\
			& DTI + DDI ($\lambda$=1.0) + PPI ($\lambda$=1.0) & \textbf{0.972 $\pm$ 0.005} & 0.971 $\pm$ 0.005\\
			& Learned-$\lambda$ & 0.969 $\pm$ 0.007 & \textbf{0.973 $\pm$ 0.005}\\
			\midrule
			\multirow{6}{*}{IC} & DTI-only & 0.973 $\pm$ 0.006 & 0.972 $\pm$ 0.005\\
			& DTI + DDI & 0.661 $\pm$ 0.158 & 0.971 $\pm$ 0.006\\
			& DTI + PPI & 0.969 $\pm$ 0.007 & \textbf{0.975 $\pm$ 0.006}\\
			& DTI + DDI ($\lambda$=0.5) + PPI ($\lambda$=0.5) & \textbf{0.975 $\pm$ 0.003} & 0.973 $\pm$ 0.004\\
			& DTI + DDI ($\lambda$=1.0) + PPI ($\lambda$=1.0) & 0.822 $\pm$ 0.181 & 0.971 $\pm$ 0.005\\
			& Learned-$\lambda$ & 0.971 $\pm$ 0.007 & 0.970 $\pm$ 0.005\\
			\midrule
			\multirow{6}{*}{DrugBank} & DTI-only & 0.903 $\pm$ 0.006 & 0.918 $\pm$ 0.004\\
			& DTI + DDI & 0.907 $\pm$ 0.005 & 0.921 $\pm$ 0.006\\
			& DTI + PPI & 0.915 $\pm$ 0.007 & 0.920 $\pm$ 0.004\\
			& DTI + DDI ($\lambda$=0.5) + PPI ($\lambda$=0.5) & 0.916 $\pm$ 0.006 & 0.923 $\pm$ 0.004\\
			& DTI + DDI ($\lambda$=1.0) + PPI ($\lambda$=1.0) & \textbf{0.921 $\pm$ 0.004} & \textbf{0.925 $\pm$ 0.004}\\
			& Learned-$\lambda$ & 0.914 $\pm$ 0.007 & 0.922 $\pm$ 0.005\\
			\bottomrule
		\end{tabular}
		
	\end{table*}
	When the fixed weight coefficient $\lambda$ is set to 0.5 for DDI and PPI, the AUC values are 0.966 and 0.972 respectively. When both DDI and PPI have full contributions, the improvement is quite stark for DTI-DM (0.972) while for DTI-CX, there is no such improvement. Moreover, when the value of $\lambda$ is learned and not fixed, AUC values do not show significant changes for DTI-DM but there is some change for DTI-CX; such $\lambda$ values are 1.072, 1.062 and 1.066 for DTI, DDI and PPI respectively for DistMult and 1.081, 1.074 and 1.073 for ComplEX. DrugBank results also show mostly similar patterns. 
	For this dataset, $\lambda$ values are 1.135 (DTI), 1.115 (DDI) and 1.119 (PPI) for DistMult and 1.131 (DTI), 1.117 (DDI) and 1.116 (PPI) for ComplEX. For IC dataset, most significant result is reported for DistMult (0.975) when DTI has full contributions and both DDI and PPI have $\lambda$ = 0.5. For ComplEX, the best result is shown when DTI and PPI are combined.
	
	The ablation study shows that incorporating auxillary relational information such as DDI and PPI can help improve the DTI prediction performance. As can be seen from the results of both Tables~\ref{tab:combined_DTI} and \ref{tab:ablation_DTI}, across all the three datasets, combining DTI, DDI and PPI in different proportions show the best results. ComplEX consistently outperforms DistMult for DTI prediction task. This improvement can be attributed in parts to its ability to model asymmetric relationships especially for drug-target interactions where the nature of the relationship is directed. Moreover, the increased expressiveness of ComplEX may allow to capture more nuanced interaction patterns beyond simple symmetric compatibility. Thus, though the performance gap remains moderate suggesting that asymmetry may have a contribution to improved modelling, it is probably not the sole determining factor. 
	
	The impact of task weighting is also evident, where equal weighting ($\lambda$ = 1.0) mostly outperforms moderate weighting ($\lambda$ = 0.5) for Enzyme and DrugBank datasets, highlighting the importance of balancing auxiliary supervision. Furthermore, the learned weighting strategy achieves competitive or superior performance, thereby demonstrating the effectiveness of adaptive task balancing. 
	
	Thus, considering both Tables~\ref{tab:combined_DTI} and Tables~\ref{tab:ablation_DTI}, it can be seen that comparison between DistMult and ComplEX does not reveal a consistent performance across all the datasets. While ComplEX is theoretically more expressive due to its ability to model asymmetric relations, this advantage is not fully realized in our setting. This can be attributed to the nature of the DTI prediction task, where interactions primarily reflect compatibility between drugs and proteins rather than strongly directional relationships. Moreover, the proposed framework already incorporates relational graph learning and attention-based feature refinement, which capture complex interaction patterns effectively. As a result, the additional expressiveness of ComplEX does not consistently translate into improved performance. These findings suggest that simpler bilinear models such as DistMult can also be sufficient for modelling interactions in the considered datasets, offering competitive performance with lower computational complexity as compared to ComplEX. 
	
	\subsubsection{Regression}
\begin{figure*}
		\centerline{
			\includegraphics[scale=0.5]{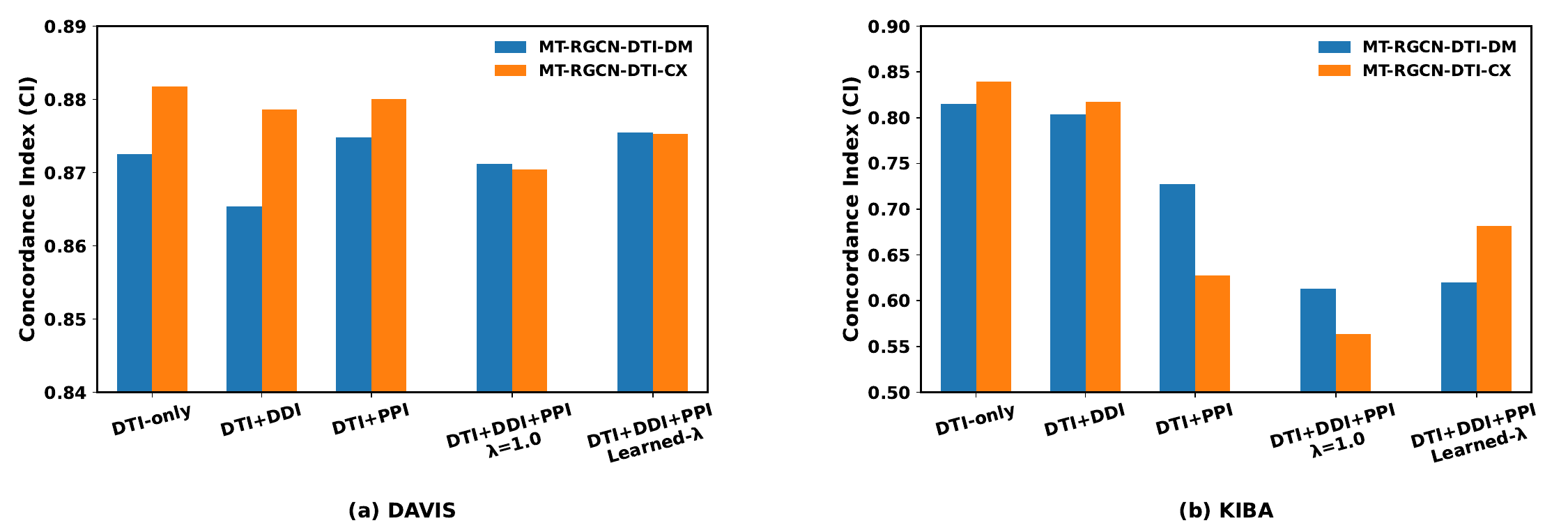}}
		\caption{Ablation Study of MT-RGCN-DTI for Regression Tasks}
		\label{Ablation_figure}
	\end{figure*}

	Figure~\ref{Ablation_figure} provides the ablation study results for regression. As can be observed from Figure~\ref{Ablation_figure}(a), both DM and CX show competent performance.
	For Davis dataset, for DTI only, the corresponding values are 0.872 $\pm$ 0.002  and 0.881 $\pm$ 0.004. Such values for DTI+DDI, DDI+PPI, DTI+DDI+PPI with $\lambda$ = 1 and learned-$\lambda$ are 0.865 $\pm$ 0.002 and 0.878 $\pm$ 0.004, 0.874 $\pm$ 0.006 and 0.880 $\pm$ 0.003, 0.871 $\pm$ 0.004 and 0.870 $\pm$ 0.006, 0.875 $\pm$ 0.004 and 0.875 $\pm$ 0.006 respectively. For KIBA dataset, for DTI, DTI+DDI, DDI+PPI, DTI+DDI+PPI with $\lambda$ = 1 and learned-$\lambda$, the values for DM and CX are  0.815 $\pm$ 0.006 and  0.839 $\pm$ 0.005, 0.803 $\pm$ 0.001 and  0.817 $\pm$ 0.003, 0.727 $\pm$ 0.117 and  0.627 $\pm$ 0.145, 0.613 $\pm$ 0.133 and  0.563 $\pm$ 0.117 and  0.620 $\pm$ 0.136 and 0.681 $\pm$ 0.142 respectively.
	
	The ablation study reveals that the contribution of auxiliary DDI and PPI relations is highly dataset-dependent. On the DAVIS dataset, incorporating auxiliary relations results in only marginal changes in performance, indicating that the observed DTI graph already provides sufficient information for learning meaningful affinity representations. Among the auxiliary settings, PPI offers a slight improvement over DDI, suggesting that protein similarity provides complementary biological information, whereas drug similarity contributes comparatively less. In contrast, the KIBA dataset exhibits a different trend. The best performance is consistently achieved using only DTI relations, while the inclusion of DDI and PPI relations leads to a substantial reduction in the Concordance Index together with a significant increase in performance variance across different runs. Since the DDI and PPI networks are constructed from Tanimoto and cosine similarity rather than experimentally measured affinity values, the corresponding auxiliary objectives may conflict with the primary affinity regression task on the larger and more heterogeneous KIBA dataset. Consequently, optimizing similarity preservation and affinity prediction simultaneously can oversmooth the learned representations and introduce optimization instability. These observations suggest that the usefulness of auxiliary similarity relations depends on the extent to which the derived similarities correlate with the underlying binding affinity distribution.
	
	\section{Visualisation}
	To check that RGCN-CX places true drug-target interactions (DTIs) in a region distinct from non-interactions on unseen data, we visualize for each of the three DTI datasets (Enzyme, IC and DrugBank), the model's learned pair representation on the held-out fold-5 test pairs.
	
	\subsection{How the Visualisation is Produced}
	
	\paragraph{What is plotted}
	For every held-out fold-5 drug-target pair we extract the model's
	$128$-dimensional (interaction vector) ComplEx interaction embedding. Only fold-5 pairs are shown; these contribute no edge to the training graph, so the layout reflects generalization rather than memorized links.
	
	\paragraph{2-D projection}
	Each ComplEx interaction embedding is standardised (zero mean, unit variance per dimension) and projected to two dimensions with Principal Component Analysis (PCA). The variance captured by PC\,1 and PC\,2 is annotated on the axes. 
	
	\paragraph{Separability probe.}
	To quantify, how well the learned space separates the two
	classes, we use a linear probe: a logistic-regression classifier trained
	on top of the frozen ComplEx interaction embeddings. 
	Because the probe is
	purely linear, a single separating hyperplane with no hidden layers, it cannot
	itself disentangle complex structure, so a high score is attributable to the
	quality of the representation rather than to the capacity of the probe. The probe
	is fitted only on the training-fold embeddings and evaluated on the
	held-out fold-5 embeddings, which are seen neither during model training nor during probe fitting. 
	We use class-balanced weights so that the heavily imbalanced IC dataset is not
	dominated by the majority (non-interaction) class, and report the
	area under the ROC curve (AUROC). The AUROC is computed on the full
	$128$-dimensional interaction embedding, whereas the scatter shows only its first two
	principal components, so the probe certifies separability that the 2-D projection
	alone cannot display.
	
	\paragraph{Figure elements}
	Orange markers are true DTIs and blue markers are non-interactions. Shaded contours are a kernel-density estimate (KDE) of the positive class; the strips above and to the right of each panel are class-conditional marginal KDEs of PC\,1 and PC\,2, respectively.
	
	\subsection{Results and Interpretation}
	
	Table~\ref{tab:viz-summary} summarises the held-out fold-5 composition, the linear-probe AUROC, and the variance captured by the two plotted components.
	Across all three datasets the held-out true interactions occupy a coherent
	region displaced from the non-interaction cloud, and the linear-probe AUROC
	($0.909$--$0.981$) confirms that this separation generalizes to unseen pairs
	rather than being an artefact of the projection.
	
	\begin{table}[t]
		\centering
		\caption{Held-out fold-5 evaluation underlying Fig.~\ref{fig:embedding-viz}.
			AUROC is from a class-balanced logistic-regression linear probe on the
			$128$-d ComplEx interaction embedding $\phi$ (trained on training folds,
			evaluated on fold~5). PC\,1/PC\,2 variance refers to the two PCA components
			shown in the scatter plots.}
		\label{tab:viz-summary}
		\begin{tabular}{lrrrr}
			\toprule
			Dataset  & Held-out pairs & Positive /  & Linear-probe  & PC\,1 / PC\,2 var.\  \\
			& & Negative & AUROC & (\%) \\
			\midrule
			Enzyme   & $1{,}170$ & $585 / 585$         & $0.981$ & $14.5 / 9.9$ \\
			IC       & $8{,}568$ & $294 / 8{,}274$     & $0.980$ & $10.2 / 7.3$ \\
			DrugBank & $2{,}334$ & $1{,}167 / 1{,}167$ & $0.909$ & $17.4 / 11.2$ \\
			\bottomrule
		\end{tabular}
	\end{table}
	\begin{figure*}
		\centerline{
			\includegraphics[height=2.5in,width=2.5in]{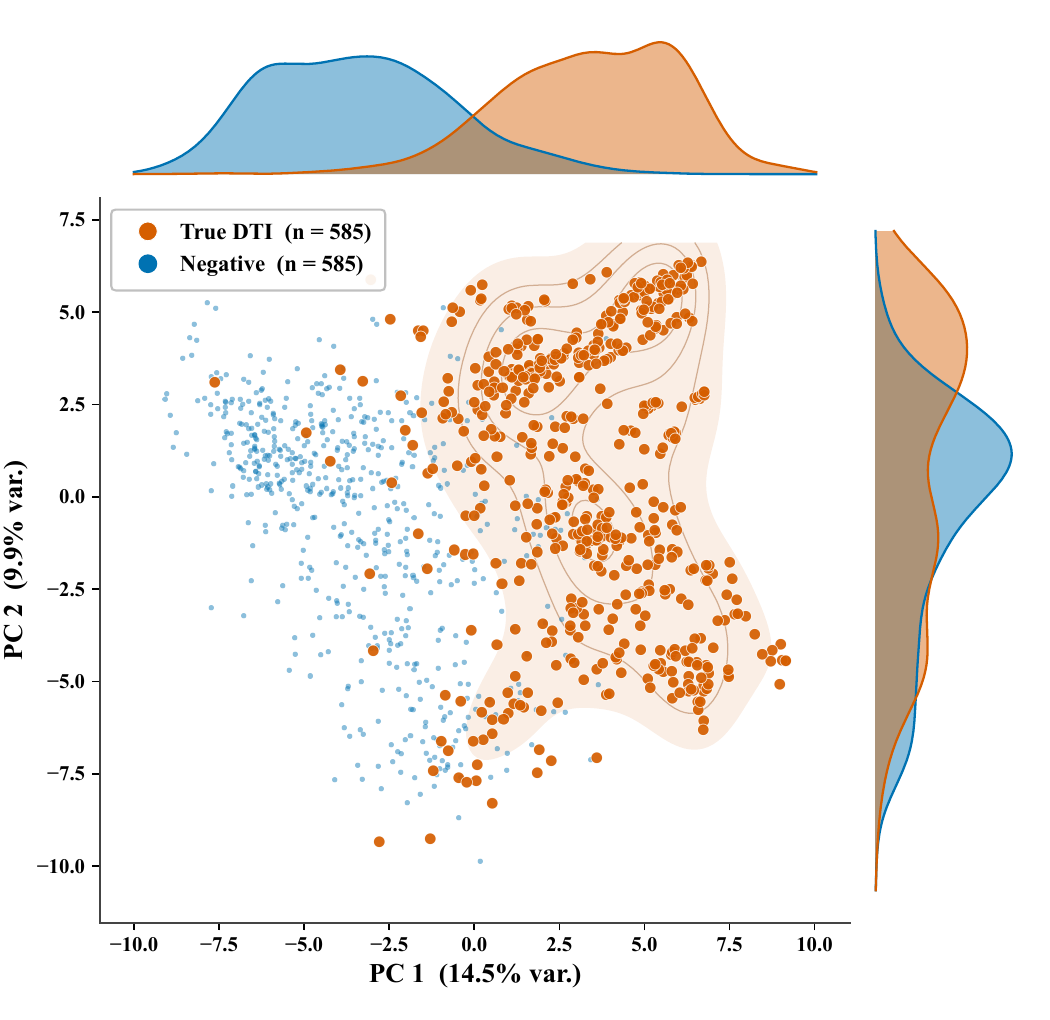}
			\includegraphics[height=2.5in,width=2.5in]{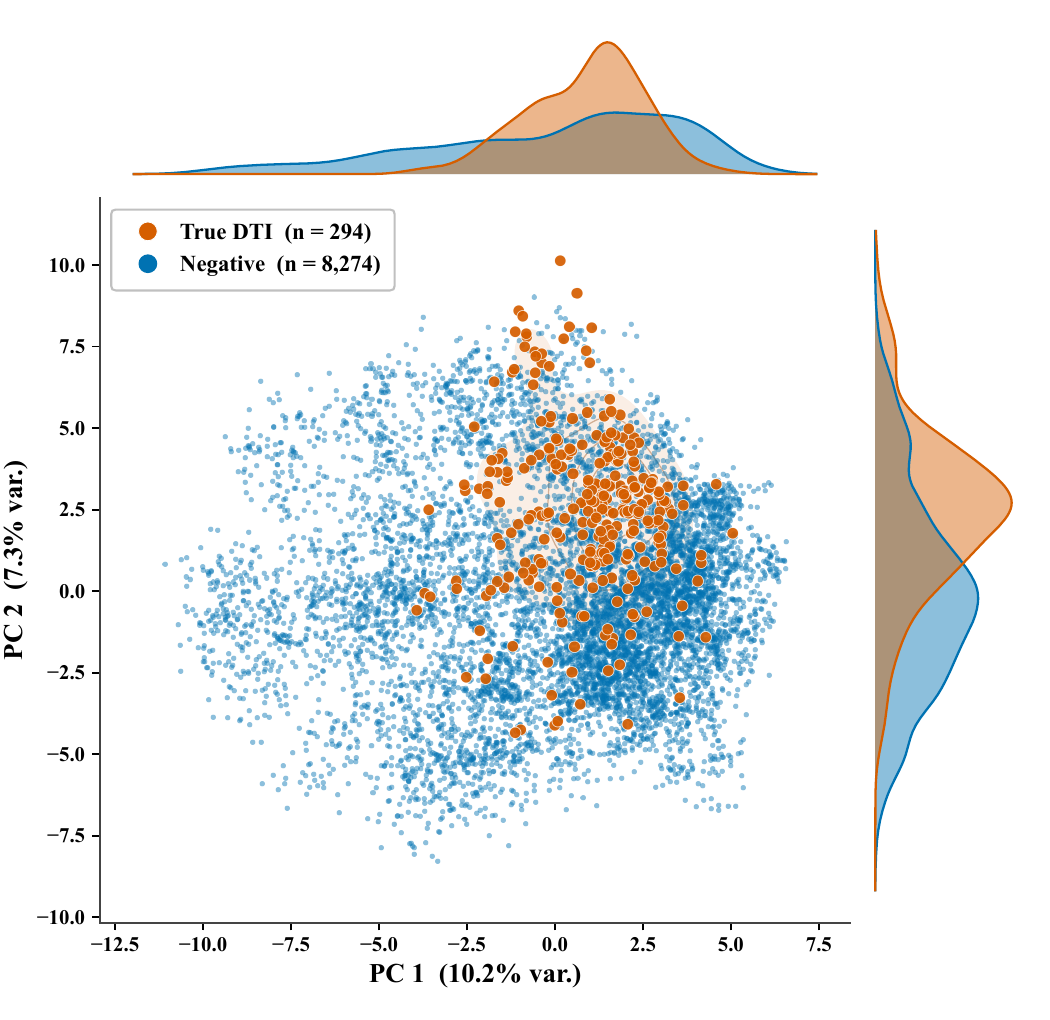}
			\includegraphics[height=2.5in,width=2.5in]{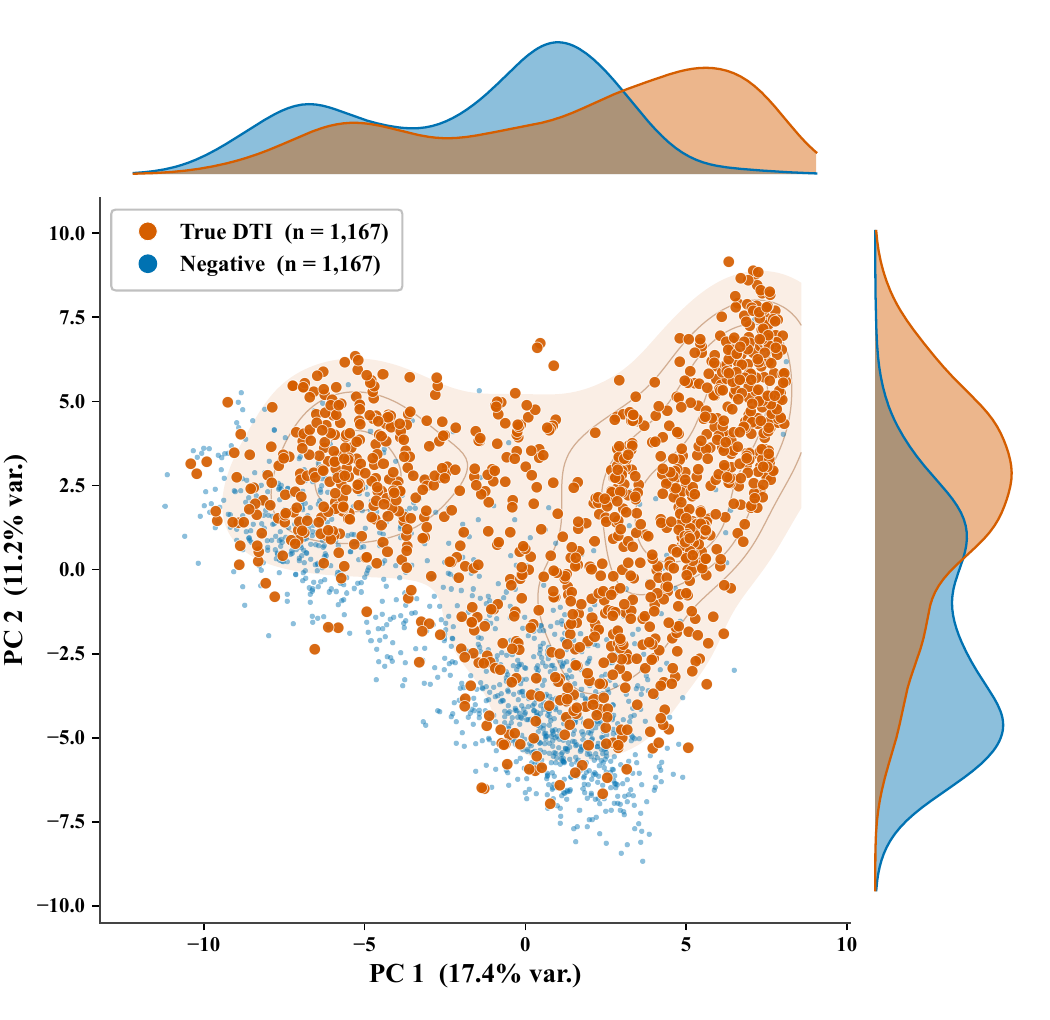}
		}
		
		\centerline{(a)\hspace{60mm}(b)\hspace{60mm}(c)}
		\caption{PCA projection of the held-out fold-5 ComplEx interaction embeddings
			for RGCN-CX on the (a) Enzyme, (b)~IC, and (c)~DrugBank datasets. Each
			point is one held-out drug--target pair: orange $=$ true DTI, blue $=$
			non-interaction. Shaded contours are the KDE of the positive class; the top
			and right strips are class-conditional marginal KDEs of PC\,1 and PC\,2.}
		\label{fig:embedding-viz}
	\end{figure*}
	\paragraph{Enzyme}
	The Enzyme fold is balanced ($585$ each) and shows the cleanest geometry:
	positives lie on the right, non-interactions on the left, with nearly disjoint
	PC\,1 marginals. The probe reaches the highest AUROC, $0.981$, i.e.\ the held-out
	classes are almost linearly separable in the learned space.
	
	\paragraph{IC}
	The IC test fold is strongly imbalanced ($294$ positives vs.\ $8{,}274$
	negatives, $\approx 3.4\%$ positive). The true interactions still form a compact
	cluster within the much larger negative cloud, and the probe reaches AUROC
	$=0.980$ even though PC\,1 and PC\,2 capture only $17.5\%$ of the variance, the
	discriminative signal is spread across many dimensions, so the 2-D view
	understates the true separability.
	
	\paragraph{DrugBank}
	The DrugBank fold is balanced ($1{,}167$ each for positive and negative). Positives occupy a broad upper
	region while non-interactions shift toward the lower-left, a clear but
	partially overlapping separation, the most overlapping of the three and
	consistent with the lowest (yet still strong) AUROC of $0.909$.
	\section{Conclusion}
	This work proposes MT-RGCN-DTI, an end-to-end relational graph convolutional network for drug–target interaction prediction. The proposed framework constructs a heterogeneous information network by integrating drug–target interactions with drug–drug and protein–protein similarity relations, enabling relation-aware representation learning through RGCN. Furthermore, knowledge graph embedding decoders, namely DistMult and ComplEx, are employed to model different biological relations within a unified framework. Unlike conventional approaches that focus solely on classification, the proposed framework supports both DTI classification and DTA regression through a unified multi-task architecture.
	
	Experimental results on multiple benchmark datasets demonstrate that MT-RGCN-DTI achieves competitive performance for DTI classification and strong performance for DTA regression compared with the state-of-the-art HIN-MTDTI framework. The ablation studies further reveal that the effectiveness of auxiliary similarity relations is dataset-dependent, highlighting the importance of selecting appropriate auxiliary biological information for different prediction tasks.
	
	As future work, we plan to replace similarity-derived DDI and PPI networks with experimentally validated drug–drug and protein–protein interaction networks. Incorporating authentic biological interaction networks may provide richer relational information and further improve the robustness and biological relevance of the learned representations.

	\section*{Acknowledgment}
	The author acknowledges the CINECA award under the ISCRA initiative, for the availability of high-performance computing resources and support. This work is carried out under Shiv Nadar Foundation (SNF) Grant.

	\ifCLASSOPTIONcaptionsoff
	\newpage
	\fi

	
	
	\bibliographystyle{IEEEtran}
	\bibliography{ref}
	%

	%
	
	
	
	
	
	
	

\end{document}